\documentclass[aps,pre,twocolumn,superscriptaddress,showpacs]{revtex4-2}
\usepackage{array}
\usepackage{amsfonts,amssymb,amsmath,mathrsfs,multirow,rotate}
\usepackage{booktabs}
\usepackage{threeparttable}
\usepackage{epsfig}
\usepackage{threeparttable}
\usepackage{chngpage}
\usepackage{latexsym}
\usepackage{dcolumn}
\usepackage{graphics,graphicx,bm,fleqn,epic,eepic,float}
\usepackage{verbatim}
\usepackage{subfigure}
\usepackage{color}
\usepackage{xr}
\usepackage{float}
\usepackage{listings} % For including code in appendix
\usepackage{placeins} % For putting plots in the desired location
\usepackage{tikz}
\usepackage{url}
\usepackage{soul}

\definecolor{red}{rgb}{1,0,0}

\definecolor{blue}{rgb}{0,0,1}

\definecolor{green}{rgb}{0,0.7,0}

\definecolor{lightgreen}{rgb}{0,0.85,0.35}

\begin{document}

\title{Model-free tracking control of complex dynamical trajectories with machine learning}
\date{\today}

\author{Zheng-Meng Zhai}
\affiliation{School of Electrical, Computer and Energy Engineering, Arizona State University, Tempe, AZ 85287, USA}

\author{Mohammadamin Moradi}
\affiliation{School of Electrical, Computer and Energy Engineering, Arizona State University, Tempe, AZ 85287, USA}

\author{Ling-Wei Kong}
\affiliation{School of Electrical, Computer and Energy Engineering, Arizona State University, Tempe, AZ 85287, USA}

\author{Bryan Glaz}
\affiliation{Vehicle Technology Directorate, CCDC Army Research Laboratory, 2800 Powder Mill Road, Adelphi, MD 20783-1138, USA}

\author{Mulugeta Haile}
\affiliation{Vehicle Technology Directorate, CCDC Army Research Laboratory, 6340 Rodman Road, Aberdeen Proving Ground, MD 21005-5069, USA}

\author{Ying-Cheng Lai} \email{Ying-Cheng.Lai@asu.edu}
\affiliation{School of Electrical, Computer and Energy Engineering, Arizona State University, Tempe, AZ 85287, USA}
\affiliation{Department of Physics, Arizona State University, Tempe, Arizona 85287, USA}

% The main text (not including figure legends or Mehtods) should be no more than 6,000 words.
% Avoid phrases like “novel”, “new”, “for the first time”, and “unprecedented” throughout the manuscript.
% Avoid exaggerated language like “extremely”, “outstanding”, or “amazingly” throughout the manuscript.

\begin{abstract}

%150 words
	Nonlinear tracking control enabling a dynamical system to track a desired trajectory is fundamental to robotics, serving a wide range of civil and defense applications. In control engineering, designing tracking control requires complete knowledge of the system model and equations. We develop a model-free, machine-learning framework to control a two-arm robotic manipulator using only partially observed states, where the controller is realized by reservoir computing. Stochastic input is exploited for training, which consists of the observed partial state vector as the first and its immediate future as the second component so that the neural machine regards the latter as the future state of the former. In the testing (deployment) phase, the immediate-future component is replaced by the desired observational vector from the reference trajectory. We demonstrate the effectiveness of the control framework using a variety of periodic and chaotic signals, and establish its robustness against measurement noise, disturbances, and uncertainties.

\end{abstract}

\maketitle

\section*{Introduction} \label{sec:intro}

The traditional field of controlling chaotic dynamical systems mostly deals 
with the problem of utilizing small perturbations to transform a chaotic 
trajectory into a desired periodic one~{\cite{OGY:1990}}. The basic principle 
is that the dynamically invariant set that generates chaotic motions contains
an infinite number of unstable periodic orbits. For any desired system 
performance, it is often possible to find an unstable periodic orbit whose
motion would produce the required behavior. The problem then becomes one to
stabilize the system's state-space or phase-space trajectory around the desired 
unstable periodic orbit, which can be achieved through linear control in the
vicinity of the orbit, thereby requiring only small control perturbations. The 
control actions can be calculated from the locations and the eigenvalues of the
target orbit, which are often experimentally accessible through a measured time
series, without the need to know the actual system 
equations~\cite{OGY:1990,GL:1997a,GL:1997b,BGLMM:2000}. Controlling chaos can 
thus be done in a model-free, entirely data-driven manner, and the control is 
most effective when the chaotic behavior is generated by a low-dimensional 
invariant set, e.g., one with one unstable dimension or one positive Lyapunov
exponent. However, for high-dimensional dynamical systems, controlling complex 
nonlinear dynamical networks is an active area of 
research~\cite{ZYA:2017,KSS:2017,JL:2019a}.

The goal of tracking control is to design a control law to enable the output of
a dynamical system (or a process) to track a given reference signal. For linear
feedback systems, tracking control can be mathematically designed with rigorous
guarantee of stability~\cite{AM:book}. However, nonlinear tracking control is 
more challenging, especially when the goal is to make a system to track a 
complex signal. In robotics, for instance, a problem is to design control 
actions to make the tip of a robotic arm, or the end effector, to follow a 
complicated or chaotic trajectory. In control engineering, designing tracking 
control typically requires complete knowledge of the system model and equations.
Existing methods for this include feedback linearization~\cite{CHARLET1989143}, 
back-stepping control~\cite{317980}, Lyapunov redesign~\cite{4790643}, and 
sliding mode control~\cite{FURUTA1990145}. These classic nonlinear control 
methods may face significant challenges when dealing with high-dimensional 
states, strong nonlinearity or time delays~\cite{2014JFM,Barros2014}, 
especially when the system model is inaccurate or unavailable. Developing 
model-free and purely data-driven nonlinear control methods is 
thus at the forefront of research. In principle, data-driven control has the 
advantage that the controller is able to adjust in real-time to new dynamics 
under uncertain conditions, but existing controllers are often not sufficiently
fast ``learners'' to accommodate quick changes in the system dynamics or 
control objectives~\cite{duriez2017machine}. In this regard, tracking a complex
or chaotic trajectory requires that the controller be a ``fast responder'' as 
the target state can change rapidly. At the present, developing model-free and 
fully data-driven control for fast tracking of arbitrary trajectories, whether 
simple or complex (ordered or chaotic), remains to be an challenging 
problem. This paper aims to address this challenge by leveraging recent 
advances in machine learning. 

Recent years have witnessed a rapid expansion of machine learning with 
transformative impacts across science and engineering. This progress has been 
fueled by the availability of vast quantities of data in many fields as well as
by the commercial success in technology and marketing~\cite{duriez2017machine}.
In general, machine learning is designed to generate models of a system from 
data. Machine-learning control is of particular relevance to our work, where 
a machine-learning algorithm is applied to control a complex system and 
generate an effective control law that maps the desired system output to the 
input. More specifically, for complex control problems where an accurate model 
of the system is not available, machine learning can leverage the experience 
and data to generate an effective controller. Earlier works on machine-learning
control concentrated on discrete-time systems, but the past few years have seen 
growing efforts in incorporating machine learning into control theory for
continuous-time systems in various 
applications~\cite{weinan2017,2020arXiv200605604B,ma2020,recht2018tour}. 

There are four types of problems associated with machine-learning control: 
control parameter identification, regression based control design of the first
kind, regression based control design of the second kind, and reinforcement 
learning. For control parameter identification, the structure of the control 
law is given but the parameters are unknown, an example of which is developing 
genetic algorithms for optimizing the coefficients of a classical controller 
[e.g., PID (proportional-integral-derivative) control or discrete-time optimal 
control~\cite{xu2019generalizable,rajalakshmi2022machine}]. For 
regression-based control design of the first kind, the task is to use machine 
learning to generate an approximate nonlinear mapping from sensor signals to 
actuation commands, an example of which is neural-network enabled computation 
of sensor feedback from a known full state feedback~\cite{pradeep2016nonlinear}.
For regression-based control design of the second kind, machine learning is 
exploited to identify arbitrary nonlinear control laws that minimize the cost 
function of the system. In this case, it is not necessary to know the model, 
control law structure, or the optimizing actuation command, and optimization is
solely based on the measured control performance (cost function), for which 
genetic programming represents an effective regression 
technique~\cite{diveev2021machine,shmalko2021control}. For reinforcement 
learning, the control law can be continually updated over measured performance 
changes based on rewards~\cite{razavi2022adaptive,waltz1965heuristic,adam2011experience,moradi2022defending,qi2019deep,henze2003evaluation,liu2006experimental,kretchmar2001robust}. It should be noted that historically, reinforcement 
learning control is not always model free. For instance, an early 
work~\cite{doya2002multiple} proposed a model-based learning method for 
nonlinear control where the basic idea is to decompose a complex task into 
multiple domains in space and time based on the predictability of the dynamics
of the environment. A framework was developed~\cite{MODARES20141780,6787009} 
to determine both the feedback and feed-forward components of the control input
simultaneously, enabling reinforcement learning to solve the tracking problem 
without requiring complete knowledge of the system dynamics and leading to the 
on- and off-policy algorithms~\cite{8169685}.

Since our aim is to achieve tracking control of complex and chaotic 
trajectories, a natural choice of the machine-learning framework is reservoir 
computing~\cite{Jaeger:2001,MNM:2002,ASSDMDSMF:2011} that has been demonstrated
to be powerful for model-free prediction of nonlinear and chaotic systems~\cite{LPHGBO:2017,PLHGO:2017,PHGLO:2018,TYHNKTNNH:2019,JL:2019b,FJZWL:2020,Bollt:2021,GBGB:2021,KFGL:2021a,FKLW:2021,KLNPB:2021,KFGL:2021b,XKSL:2021,PCGPO:2021}.
The core of reservoir computing is recurrent neural network (RNN) with low 
training cost where regularized linear regression is sufficient for training.  
Reservoir computing, shortly after its invention, was exploited to control 
dynamical systems~\cite{Jaeger:2008} where an inverse model was trained to map 
the present state and the desired state of the system to the control signal  
(action). Subsequently, the trained reservoir computer was exploited as a 
model-free nonlinear feedback controller~\cite{WWS:2012} as well as for 
detecting unstable periodic orbits and stabilizing the system about a desired 
orbit~\cite{ZML:2019}. Reservoir computing and its variant echo state Gaussian 
process~\cite{CD:2011} were also used in model predictive control of unknown 
nonlinear dynamical systems~\cite{PW:2012,HCXZ:2019}, which served as 
replacements of the traditional recurrent neural-network models with low 
computational cost. More recently, deep reservoir networks were proposed for 
controlling chaotic systems~\cite{canaday2021model}. 

%{\color{blue}
In this paper, we tackle the challenge of model-free and data-driven nonlinear 
tracking of various reference trajectories, including complex chaotic 
trajectories, with an emphasis on their potential applications in robotics. 
In particular, we examine the case of a two-arm robotic manipulator with the 
control objective of tracking any trajectories while using only partially
observed states, denoted as vector $\mathbf{y}(t)$. 
Our control framework has the following three features: (1) requirement of
only partial state observation for both training and testing, (2) a 
machine-learning training scheme that involves the observed vectors at two
consecutive time steps: $\mathbf{y}(t)$ and $\mathbf{y}(t+dt)$, and
(3) use of a stochastic signal as the input control signal for training.
With respect to feature (1), it may be speculated that the classical Takens
delay-coordinate embedding methodology could be used to construct the full
phase space from partial observation. However, in this case, the reconstructed
state is equivalent to the original system but only in a topological sense:
there is no exact state correspondence between the reconstructed and the 
original dynamical systems. For reservoir-computing based prediction and 
control tasks, such an exact correspondence is required. To our 
knowledge, achieving tracking control based on partial state observation is 
novel. In terms of features (2) and (3), we note a previous 
work~\cite{WWS:2012} on machine-learning stabilization of linear and 
low-dimensional nonlinear dynamical systems, where the phase-space region to 
realize control is localized. This was effectively an online learning approach.
In general, online learning algorithms have difficulties such as instability, 
modeling complexity as required for nonlinear control, and computational 
efficiency. For example, it is difficult for online learning to capture the 
intricate complex nonlinear dynamics, causing instability during control. 
Trajectory divergence is another common problem associated with online 
learning control, where sudden and extreme changes in the state can occur. In 
fact, as the dimension and complexity of the system to be controlled increase, 
online learning algorithms tend to fail. In contrast, offline learning is 
computationally extremely efficient and allows for more comprehensive and 
complex model training with minimum risk of trajectory divergence through 
repeated training. Our tracking framework entails following a dynamic and 
time-varying (even chaotic) trajectory in the whole phase space, where the 
offline controller can not only respond to disturbances and system variations 
but also adjust the control inputs to make the system output follow a 
continuously changing reference signal. As we will demonstrate, our control 
scheme brings these features together to enable continuous tracking of 
arbitrary complex trajectories.
% }

\section*{Results} \label{sec:results}

\begin{figure*}[ht!]
\centering
\includegraphics[width=0.8\linewidth]{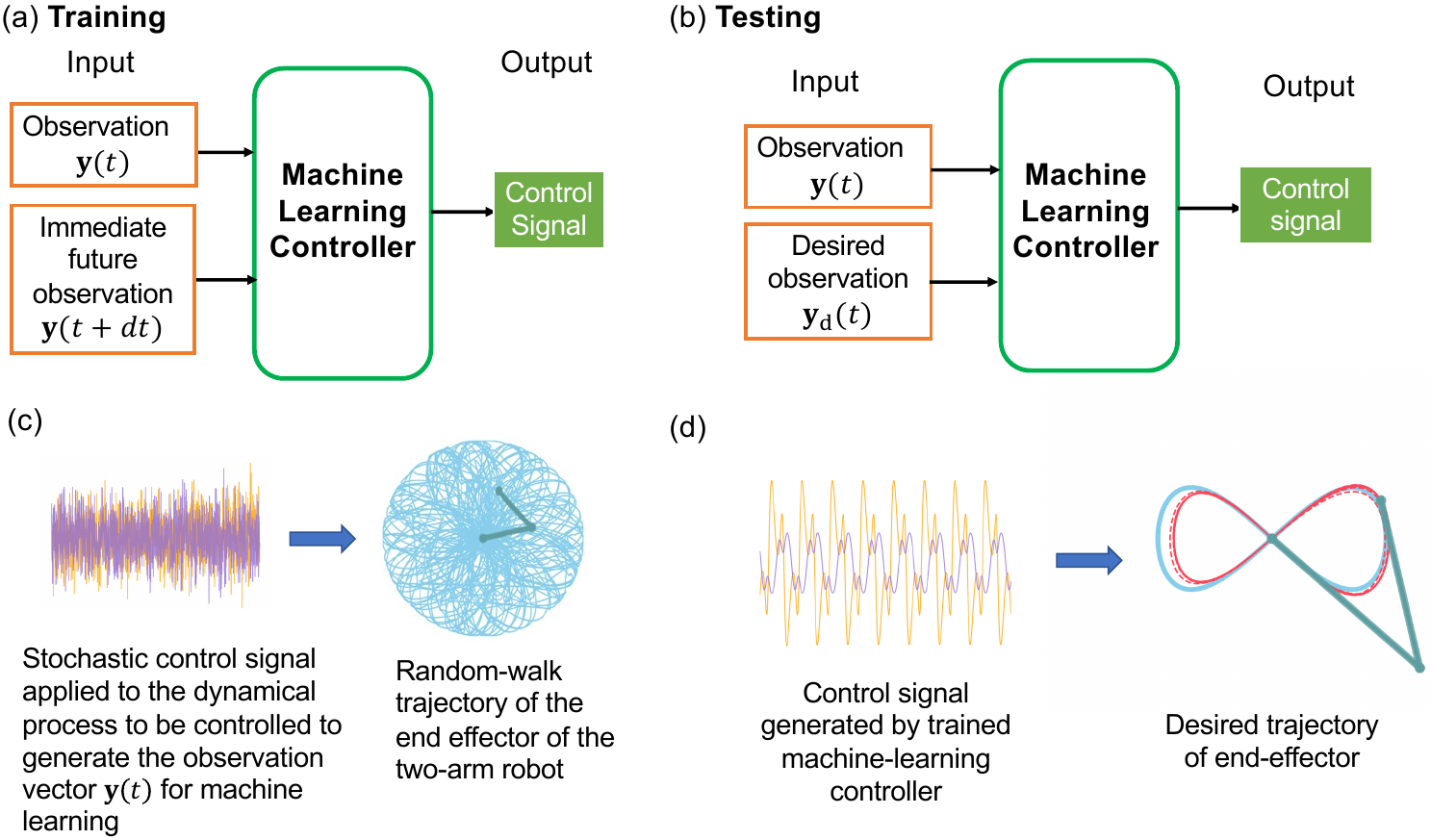}
\caption{Illustration of our proposed machine-learning tracking controller.
(a) During the training phase, the input to the machine-learning controller
consists of two vectors of equal dimension: the partial observation vector
$\mathbf{y}(t)$ and its immediate-future counterpart $\mathbf{y}(t+dt)$ as the 
complementary. The output is a control signal which, when applied to the 
dynamical system or process, will enable it to track any desired reference 
trajectory. This input configuration stipulates that the complementary 
component of the input is the immediate future state of the observation vector.
(b) In the testing phase, the complementary component of the input vector is 
replaced by $\mathbf{y}_{\rm d}(t)$, the observation vector calculated from the 
reference trajectory. Since the machine-learning controller has been trained 
to recognize the complementary input component as the immediate future state, 
in the testing phase the controller will ``force'' the observation vector to 
follow the desired vector, thereby realizing accurate tracking. Note that 
$\mathbf{y}_{\rm d}(t)$ is provided to the machine-learning controller 
according to the desired trajectory, so no process model is required. 
(c) A fully stochastic control signal is used for training, which generates 
a random-walk type of reference trajectory. The required input vectors 
$\mathbf{y}(t)$ and $\mathbf{y}(t+dt)$ to machine learning are obtained by 
observing the dynamical process to be controlled, so a mathematical model of 
the process is not required. (d) A well-trained machine learning controller 
generates the appropriate control signal to track any desired trajectory, 
where the blue and dotted red traces correspond to the reference and tracked
trajectories, respectively.}
\label{fig:new}
\end{figure*}

A more detailed explanation of the three features and their combination to 
solve the complex trajectory tracking problem is as follows.
First, existing works on reservoir-computing based controllers relied on full 
state measurements~\cite{Jaeger:2008,WWS:2012,ZML:2019,PW:2012,HCXZ:2019,canaday2021model}, but our controller requires measuring only a partial set of the 
state variables. Second, as shown in Fig.~\ref{fig:new}(a), during the 
training phase, the input to the machine learning controller consists of two 
components: the observation vector at two consecutive time steps: 
$\mathbf{y}(t)$ and $\mathbf{y}(t+dt)$. That is, at any time step $t$,
the second vector is the state of the observation vector in the immediate 
future. This input configuration offers several advantages, which are evident 
in the testing phase, as shown in Fig.~\ref{fig:new}(b). After the
machine-learning controller has been trained, the testing input consists of 
the observation vector $\mathbf{y}(t)$ and the desired observation vector 
$\mathbf{y}_{\rm d}(t)$, calculated from the reference trajectory to be 
tracked. The idea is that, during the testing or deployment, the immediate 
future state of the observation is manipulated to match the desired vector 
from the trajectory. This way, the output control signal from the 
machine-learning controller will make the end effector of the robotic 
manipulator to precisely trace out the desired reference trajectory. The third 
feature is the choice of the control signal for training. Taking advantage of 
the fundamental randomness underlying any chaotic trajectory, we conduct the 
training via a completely stochastic control input, as shown in 
Fig.~\ref{fig:new}(c), where the reference trajectory generated by such a 
control signal through the underlying dynamical process is a random walk. 
Compared with a deterministic chaotic trajectory with short-term 
predictability, the random-walk trajectory is more complex as its movements 
are completely unpredictable. As a result, the machine-learning controller 
trained with a stochastic signal will possess a level of complexity sufficient 
for controlling or overpowering any deterministic chaotic trajectory. 
In general, our machine-learning controller so trained is able to learn a 
mapping between the state error and a suitable control signal for any 
reference trajectory. In the testing phase, given the current and desired 
states, the machine-learning controller generates the control signal that 
enables the robotic manipulator to track any desired complex reference 
trajectory, as illustrated in Fig.~\ref{fig:new}(d). We demonstrate the 
working and power of our machine-learning tracking control using a variety of 
periodic and chaotic trajectories, and establish the robustness against 
measurement noise, disturbances, and uncertainties. While our primary 
machine-learning scheme is reservoir computing, we also test the 
architecture of feed-forward neural networks and demonstrate its working as an 
effective tracking controller, albeit with higher computational time complexity.
Overall, our work provides a powerful model-free data-driven control framework 
that only relies on partial state observation and can successfully track 
complex or chaotic trajectories. 

\subsection*{Principle of machine-learning based control}

\begin{figure}[ht!]
\centering
\includegraphics[width=\linewidth]{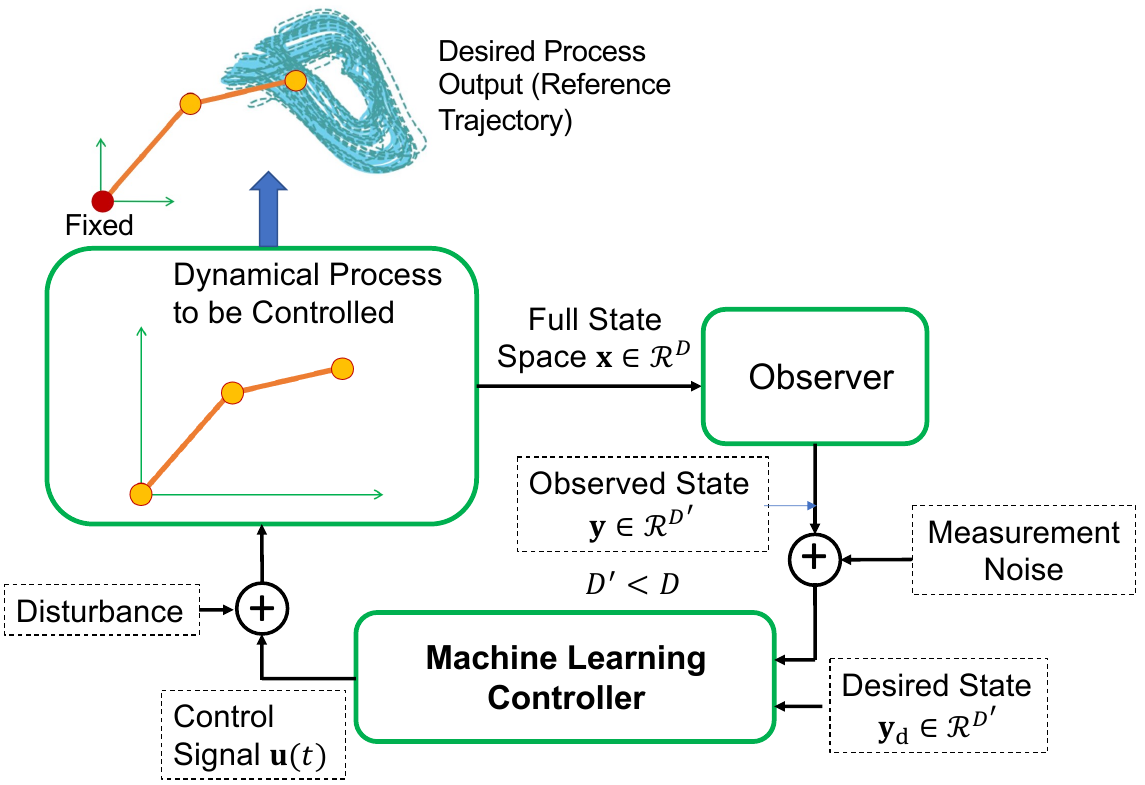}
\caption{Working principle of our machine-learning based tracking control.
The state space vector $\mathbf{x}$ of the dynamical process to be controlled 
is $D$-dimensional. An observer produces a $D'$-dimensional measurement vector 
$\mathbf{y}$, where $D'<D$. The machine-learning controller uses this vector 
and the corresponding desired vector $\mathbf{y}_{\rm d}$ calculated from the 
reference trajectory to be tracked as the input and generates a proper, 
typically lower-dimensional control signal $\mathbf{u}(t)$. Disturbance
is applied to the control signal vector $\mathbf{u}$ and measurement noise is 
present during the observation of the state vector $\mathbf{x}$. Unlike 
controllers that rely on the error between $\mathbf{y}$ and $\mathbf{y}_{\rm d}$, 
our controller uses both signals as inputs, which provides it with two degrees 
of freedom.}
\label{fig:overview}
\end{figure}

An overview of the working principle of our machine-learning based tracking 
control is as follows. Consider a dynamical process to be controlled, e.g., a 
two-arm robotic system, as indicated in the green box on the left in 
Fig.~\ref{fig:overview}. The objective of control is to make the end effector, 
which is located at the tip of the outer arm, track a complex trajectory. Let 
$\mathbf{x} \in \mathcal{R}^D$ represent the full, $D$-dimensional state space 
of the process. An observer has access to part of the full state space and 
produces a $D'$-dimensional measurement vector $\mathbf{y}$, where $D' < D$.
A properly selected and trained machine-learning scheme takes $\mathbf{y}$ as
its input and generates a low-dimensional control signal 
$\mathbf{u}(t) \in \mathcal{R}^{D''}$ (e.g., two respective torques applied 
to the two arms), where $D'' \le D'$, to achieve the control objective. 
The workings of our control scheme can be understood in terms of the following 
three essential components: (1) a mathematical description of the dynamical 
process and the observables ({\bf Methods}), (2) a physical description of how 
to obtain the control signals from the observables (known as inverse dynamics -
{\bf Methods}) and (3) the machine-learning scheme (Supplementary Note 1). 

\begin{figure}[ht!]
\centering
\includegraphics[width=\linewidth]{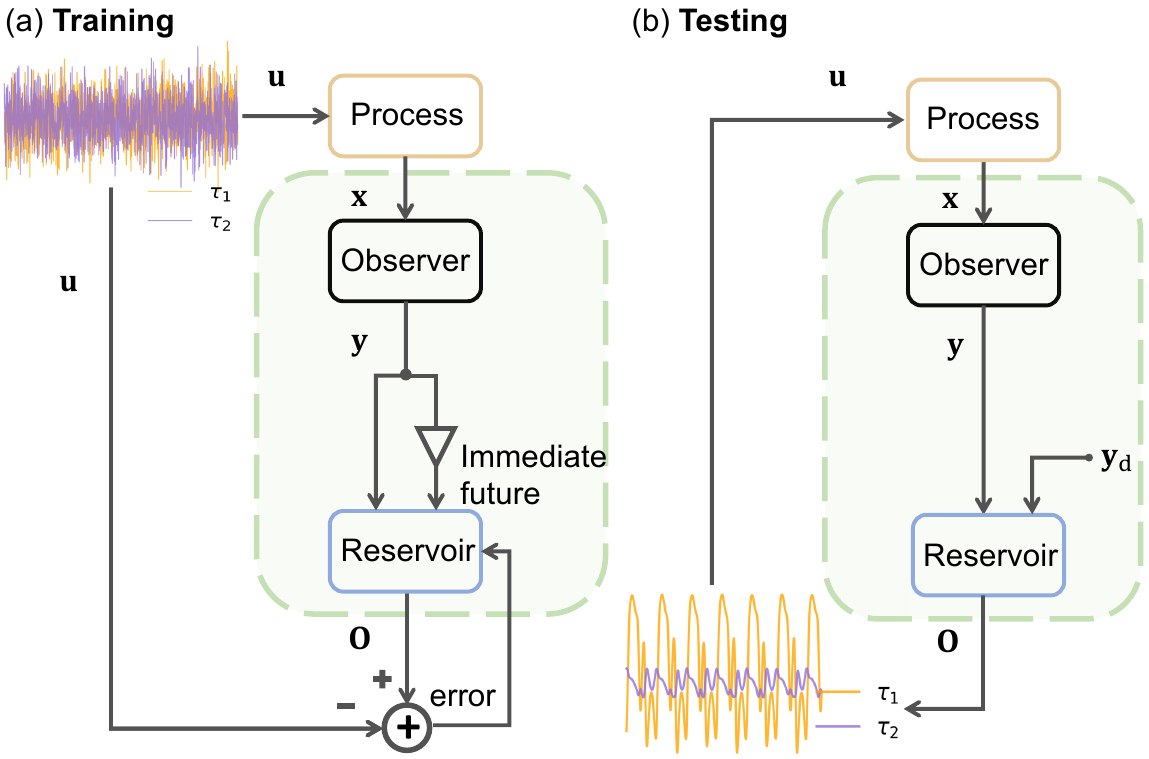} 
\caption{Basic architecture of proposed machine-learning based tracking control
framework for the two-arm robotic system. (a) During the 
training phase, the random torques $\tau_1(t)$ and $\tau_2(t)$ are used as the 
control signal to the dynamical process (the two-arm system) to generate the 
state vector $\mathbf{x}(t)$. An observer produces a lower-dimensional observed
vector $\mathbf{y}(t)$. This vector and its immediate future  
$\mathbf{y}(t+dt)$ are used as the input to the reservoir computing machine, 
whose output is a two-dimensional torque vector. The difference between the 
reservoir output and the original random torque signal constitutes the error 
for training. (b) In the testing or deployment phase, the input to the 
reservoir computer is the observed vector $\mathbf{y}(t)$ and the desired 
vector $\mathbf{y}_{\rm d}(t)$ calculated from the reference trajectory. The output 
of the reservoir is a control signal that drives the two-arm system so that 
its end effector precisely traces out the desired reference trajectory.}
\label{fig:ML}
\end{figure}

The state variable of the two-joint robot-arm system is eight-dimensional:
$\mathbf{x}\equiv [C_x,C_y,q_1,q_2,\dot{q}_1,\dot{q}_2,\ddot{q}_1,\ddot{q}_2]^T$,
where $C_x$ and $C_y$ are the Cartesian coordinates of the end effector, $q_i$, 
$\dot{q}_i$ and $\ddot{q}_i$ are the angular position, angular velocity and 
angular acceleration of aim $i$ ($i=1,2$). The measurement vector is 
four-dimensional: $\mathbf{y} \equiv [C_x, C_y, \dot{q_1}, \dot{q_2}]^T$.
A remarkable feature of our framework is that a purely stochastic signal can be 
conveniently used for training. As illustrated in Fig.~\ref{fig:new}(c), the 
torques $\tau_1(t)$ and $\tau_2(t)$ applied to the two arms, respectively, are 
taken to be stochastic signals from a uniform distribution, which produce a 
random-walk type of trajectory of the end effector. The control input for 
training is $\mathbf{u}(t) = [\tau_1(t),\tau_2(t)]^T$, as shown in 
Fig.~\ref{fig:ML}(a). To ensure continuous control input, we use a Gaussian 
filter to smooth the noise input data. With the control signal, the forward 
model Eq.~\eqref{eq:plant} (in {\bf Methods}) produces the state vector 
$\mathbf{x}(t)$ and the observer generates the vector $\mathbf{y}(t)$. The 
observed vector $\mathbf{y}(t)$ and its delayed version $\mathbf{y}(t+dt)$ 
constitute the input to the reservoir computing machine that generates a control
signal $\mathbf{O}(t)$ as the output, leading to the error signal 
$\mathbf{e}(t) = \mathbf{O}(t) - \mathbf{u}(t)$ as the loss function for 
training the neural network. 

A well trained reservoir can then be tested or deployed to generate any desired
control signal, as illustrated in Fig.~\ref{fig:ML}(b). In particular, during 
the testing phase, the input to the reservoir computer consists of the observed
vector $\mathbf{y}(t)$ and the desired vector $\mathbf{y}_{\rm d}(t)$ characterized
by the two Cartesian coordinates of the reference trajectory of the end effector
and the resulting angular velocities of the two arms. Note that, given an 
arbitrary reference trajectory $\{C_x(t),C_y(t)\}$, the two angular velocities
can be calculated (extrapolated) from Eqs.~\eqref{eq:inverse_kinematics_1} and 
\eqref{eq:inverse_kinematics_2} (in {\bf Methods}). The output of the reservoir 
computing machine is the two required torques $\tau_1(t)$ and $\tau_2(t)$ that 
drive the two-arm system so that the end effector traces out the desired 
reference trajectory.

\begin{figure}[ht!]
\centering
\includegraphics[width=\linewidth]{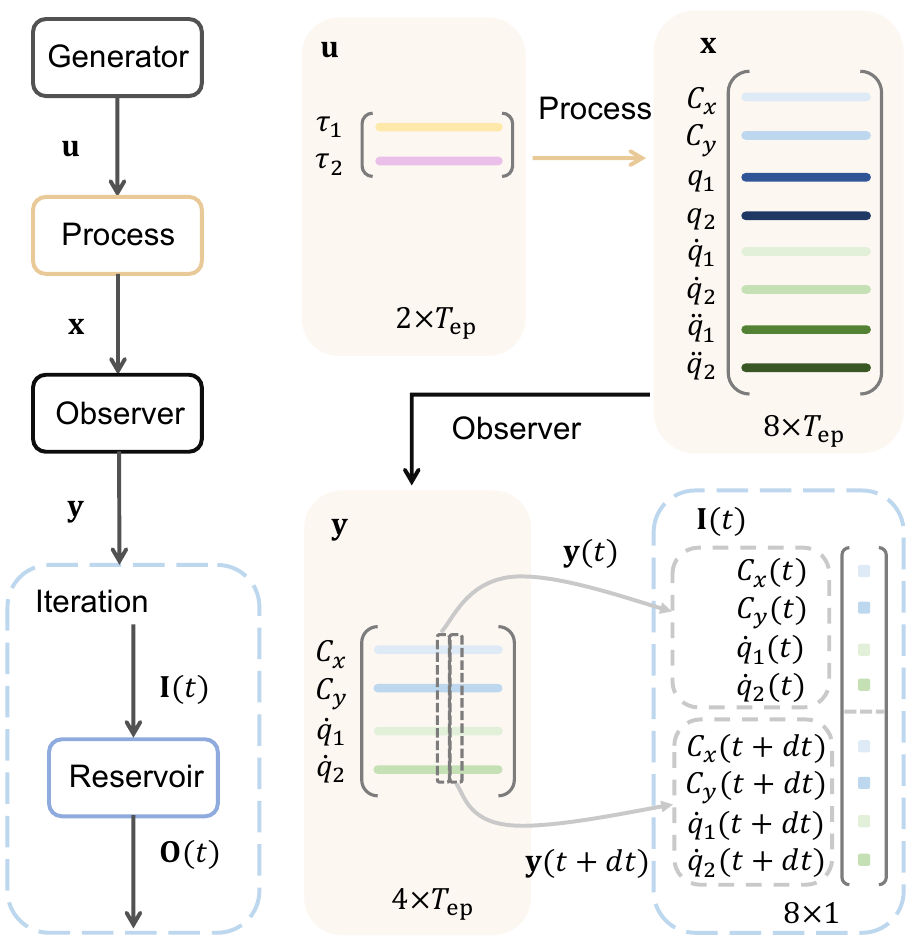}
\caption{Dynamical variables and data structure associated with the training
phase of the machine-learning controller. Two stochastic signals act as torques
for the two-arm system, causing its end effector to generate a random walk. For
a training episode of duration $T_{\rm ep}$, the input to the process is a 
$2\times T_{\rm ep}$ data matrix. The state and the observed vectors are 
represented as a $8\times T_{\rm ep}$ and a $4\times T_{\rm ep}$ matrix, 
respectively. At any time step $t$, the input to the reservoir computing machine 
is an eight-dimensional vector constituting $\mathbf{y}(t)$ and 
$\mathbf{y}(t+dt)$.}
\label{fig:ML_training}
\end{figure}

\paragraph*{Training.}
The detailed structure of the data and the dynamical variables associated with
the training process is described, as follows. The training phase is divided 
into a number of uncorrelated episodes, each of length $T_{\rm ep}$, 
which defines the resetting time. At the start of each episode, the state 
variables including $[\dot{q}_1,\dot{q}_2,\ddot{q}_1,\ddot{q}_2]$ along with 
the controller state are reset. The initial angular positions $q_1$ and $q_2$ 
are randomly chosen in their defined range, respectively. For each episode, 
the process's control input is stochastic for a time duration of $T_{\rm ep}$, 
generating a torque matrix of dimension $2\times T_{\rm ep}$, as illustrated in 
Fig.~\ref{fig:ML_training}. For the same time duration, the state $\mathbf{x}$ 
of the dynamical process and the observed state $\mathbf{y}$ can be expressed 
as a $8\times T_{\rm ep}$ and a $4\times T_{\rm ep}$ matrix, respectively. At each
time step $t$, the input to the reservoir computing machine, the concatenation
of $\mathbf{y}(t)$ and $\mathbf{y}(t+dt)$, is an $8\times 1$ vector. The neural
network learns to generate a control input that takes the process's output from
$\mathbf{y}(t)$ to $\mathbf{y}(t+dt)$ so as to satisfy the tracking goal. The 
resulting trajectory of the end effector of the process, due to the stochastic 
input torques, is essentially a random walk. To ensure that the random walk 
covers as much of the state space as possible, the training length and 
machine-learning parameters must be appropriately chosen.

\paragraph*{Testing.}
In the testing phase, the trained neural network inverts the dynamics of the 
process. In particular, given the current and desired output, the neural 
network generates the control signal to drive the system's output from 
$\mathbf{y}(t)$ to $\mathbf{y}(t+dt)$ while minimizing the error between 
$\mathbf{y}(t+dt)$ and $\mathbf{y}_{\rm d}(t+dt)$. We shall demonstrate that our 
machine-learning controller is capable of tracking any complicated 
trajectories, especially a variety of chaotic trajectories.

With a reservoir controller and the inverse model, our tracking-control 
framework is able to learn the mapping between the current and desired 
position of the end effector and deliver a proper control signal, for a given
reference trajectory. For demonstration, we use 15 different types of reference
trajectories including those from low- and high-dimensional chaotic systems. 
(The details of the generation of these reference trajectories are presented in
Supplementary Note 2) Note that the starting position of the end effector
is not on the given reference trajectory, requiring a ``bridge'' to drive the 
end effector from the starting position to the trajectory (See Supplementary
Note 3). Here we also address the issue of probability of control success and 
the robustness of our method against measurement noise, disturbance, and 
parameter uncertainties. 

\subsection*{Examples of tracking control} \label{subsec:results_1}

The basic parameter setting of the reservoir controller is as follows. The 
size of the hidden-layer network is $N_r=200$. The dimensionless time step 
of the evolution of dynamical network is $dt=0.01$. A long training length is 
chosen: $200,000/dt$ so as to ensure that the learning experience of the neural 
network extends through most of the phase space in which the reference 
trajectory resides. The testing length is $2,500/dt$, which is sufficient for 
the controller to track a good number of complete cycles of the reference 
trajectory. The values of the reservoir hyperparameters obtained through 
Bayesian optimization are: spectral radius $\rho=0.76$, input weights factor 
$\gamma=0.76$, leakage parameter $\alpha=0.84$, regularization coefficient 
$\beta=7.5\times10^{-4}$, link probability $p=0.53$, and the bias 
$w_{\rm b}=2.00$. 

\begin{figure*}[ht!]
\centering
\includegraphics[width=\linewidth]{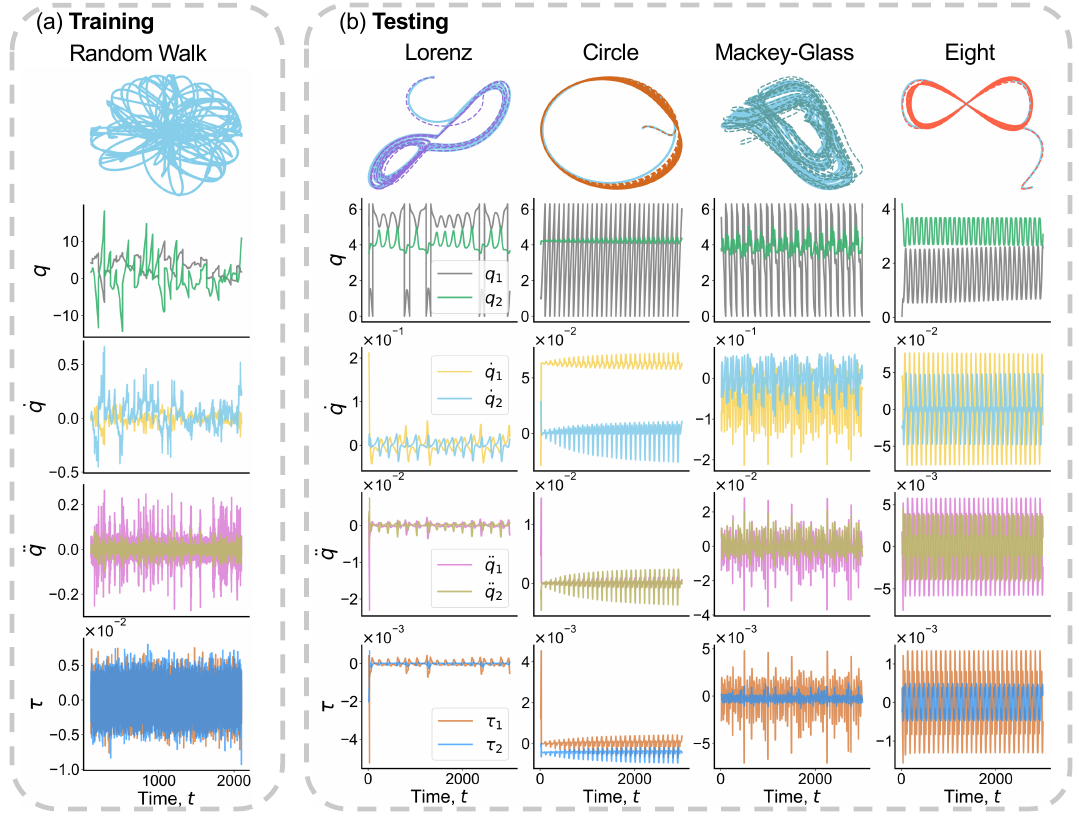} 
\caption{Examples of tracking control. (a) A random-walk reference trajectory 
used in training. The time series plots below are the angular positions ($q$), 
velocities ($\dot{q}$), and accelerations ($\ddot{q}$) as well as the two 
torques (the control signals, $\tau$) applied to the two arms. (b) Successful 
tracking of four reference trajectories: two chaotic (Lorenz and Mackey-Glass) 
and two periodic (a circle and an eight figure). Solid blue and dotted traces
represent the reference and controlled trajectories, respectively. The 
reservoir controller generates the proper control signals based on the current 
measurement vector $\mathbf{y}(t)$ and the corresponding desired vector 
$\mathbf{y}_{\rm d}(t)$.} 
\label{fig:4_traj}
\end{figure*}

The training phase is divided into a series of uncorrelated episodes, ensuring
that the velocity or acceleration of the robot arms will not become 
unreasonably large during the random-walk motion of the reference trajectory. 
The episodes are initialized at time $T_{\rm ep}=80/dt$. The angular 
positions $q_1$ and $q_2$ of the two arms is set to a random value uniformly 
distributed in the ranges $[0, 2\pi]$ and $[-\pi, \pi]$, respectively. The
angular velocities and accelerations 
$[\dot{q}_1,\dot{q_2},\ddot{q}_1,\ddot{q}_2]$ of the two arms as well as the
reservoir state $\mathbf{r}$ are set to zero initially. From the values of 
$q_1$ and $q_2$, the coordinates $C_x$ and $C_y$ of the end effector can be 
obtained from Eq.~\eqref{eq:forward}. At the beginning of each episode, since 
$q_1$ and $q_2$ are random, the end-effector will be a random point inside a 
circle of radius $l_1+l_2=1$ centered at the origin. Figure~\ref{fig:4_traj}(a)
shows the random-walk reference trajectory used in training and examples of 
the evolution of the dynamical states of the two arms (in two different colors):
$q_{1,2}(t)$, $\dot{q}_{1,2}(t)$, $\ddot{q}_{1,2}(t)$, and $\tau_{1,2}(t)$. 
To maintain the continuity of the control signal during the training phase, we 
invoke a Gaussian filter to smooth the noisy signals. Given the control signal 
$u(t)=[\tau_1(t),\tau_2(t)]$ and the state variables 
$[q_{1,2}(t),\dot{q}_{1,2}(t)]$ at each time step, the angular accelerations 
$\ddot{q}_{1,2}(t)$ can be obtained from Eq.~\eqref{eq:robot_arm}. At the next 
time step, the angular positions and velocities are calculated using
\begin{align}
	q_{1,2}(t+dt) &= q_{1,2}(t) + \dot{q}_{1,2}(t)\cdot dt, \\ \nonumber
	\dot{q}_{1,2}(t+dt)&=\dot{q}_{1,2}(t)+\ddot{q}_{1,2}(t)\cdot dt.
\end{align}
The purpose of the training is for the reservoir controller to learn the 
intrinsic mapping from $\mathbf{y}(t)$ to $\mathbf{y}(t+dt)$ and to produce an 
output control signal $u(t)=[\tau_1(t), \tau_2(t)]$. 

In the testing phase, given the current measurement $\mathbf{y}(t)$ and the 
desired measurement $\mathbf{y}_{\rm d}(t+dt)$, the reservoir controller 
generates a control signal and feed it to the process. The tracking error is 
the difference between $\mathbf{y}_{\rm d}(t+dt)$ and $\mathbf{y}(t+dt)$.
Figure~\ref{fig:4_traj}(b) presents four examples: two chaotic (Lorenz and 
Mackey-Glass) and two periodic (a circle and an eight figure) reference 
trajectories, where in each case, the angular positions, velocities, and
accelerations of both arms together with the control signal (the two torques)
delivered by the reservoir controller are shown. As the reservoir controller 
has been trained to track a random walk signal, which is fairly complex and 
chaotic, it possesses the ability to accurately track these types of 
deterministic signals. 
 
Our machine-learning controller, by design, is generalizable to arbitrarily
complex trajectories. This can be seen, as follows. 
In the training phase, no specific trajectory is used. Rather, training is
accomplished by using a stochastic control signal to generate a random-walk
type of trajectory that ``travels'' through the entire state-space domain of
interest. The machine-learning controller does not learn any specific
trajectory example but a generic map from the observed state at the current
time step to the next under a stochastic control signal. The training process
determines the parameter values for the controller, which are fixed when it
is deployed in the testing phase. The required input for testing is the
current observed state $\mathbf{y}(t)$ and the desired state $\mathbf{y}_{\rm d}(t)$
from the reference trajectory. The so-designed machine-learning controller
is capable of making the system to follow a variety of complex periodic or
chaotic trajectories to which the controller is not exposed during training.
(Supplementary Notes 2 and 4 present many additional examples.)

\subsection*{Robustness against disturbance and noise} \label{subsec:results_2}

\begin{figure}[ht!]
\centering
\includegraphics[width=\linewidth]{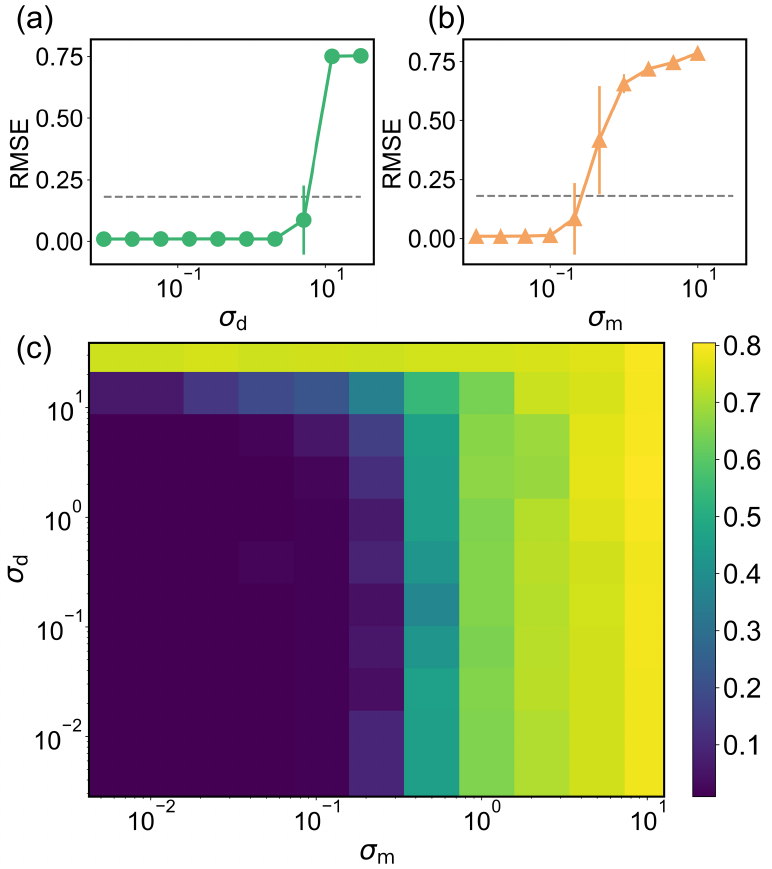} 
\caption{Robustness against disturbance and noise for tracking the chaotic
Lorenz reference trajectory. (a,b) Ensemble-averaged testing RMSE versus the 
amplitude $\sigma_{\rm d}$ of the disturbance and the noise amplitude 
$\sigma_{\rm m}$, respectively. Error bars represent standard deviation 
calculated from 50 independent realizations. In each case, the horizontal 
dashed line represents some empirical threshold below which the 
tracking-control performance may be regarded as satisfactory. The tolerance of 
tracking control to disturbance is about $\sigma_{\rm d} \alt 10^{0.5}$ and 
that to noise is about $\sigma_{\rm m} \alt 10^{-1}$. (c) Color-coded RMSE in 
the parameter plane $(\sigma_{\rm d},\sigma_{\rm m})$.}
\label{fig:noise_test}
\end{figure}

We consider normally distributed stochastic processes of zero mean and standard
deviations $\sigma_{\rm d}$ and $\sigma_{\rm m}$ to simulate disturbance and 
noise, which are applied to the control signal vector $\mathbf{u}$ and the 
process state vector $\mathbf{x}$, respectively, as shown in 
Fig.~\ref{fig:overview}. Figures~\ref{fig:noise_test}(a) and 
\ref{fig:noise_test}(b) show the ensemble-averaged testing RMSE (root mean 
square error, defined in Supplementary Note 1) versus $\sigma_{\rm d}$ and 
$\sigma_{\rm m}$, respectively, for tracking of the chaotic Lorenz reference 
trajectory, where 50 independent realizations are used to calculate the average
errors. In the case of disturbance, near zero RMSEs are achieved for 
$\sigma_{\rm d} \alt 10^{0.5}$, while the noise tolerance is about $10^{-1}$. 
Color-coded testing RMSEs in the parameter plane 
$(\sigma_{\rm d},\sigma_{\rm m})$ are shown in Fig.~\ref{fig:noise_test}(c). 
Those results indicate that, for reasonably weak disturbances and small noise, 
the tracking performance is robust. (Additional examples are presented in 
Supplementary Note 4.) 

\subsection*{Robustness against parameter uncertainties} \label{subsec:results_3}

\begin{figure}[ht!]
\centering
\includegraphics[width=\linewidth]{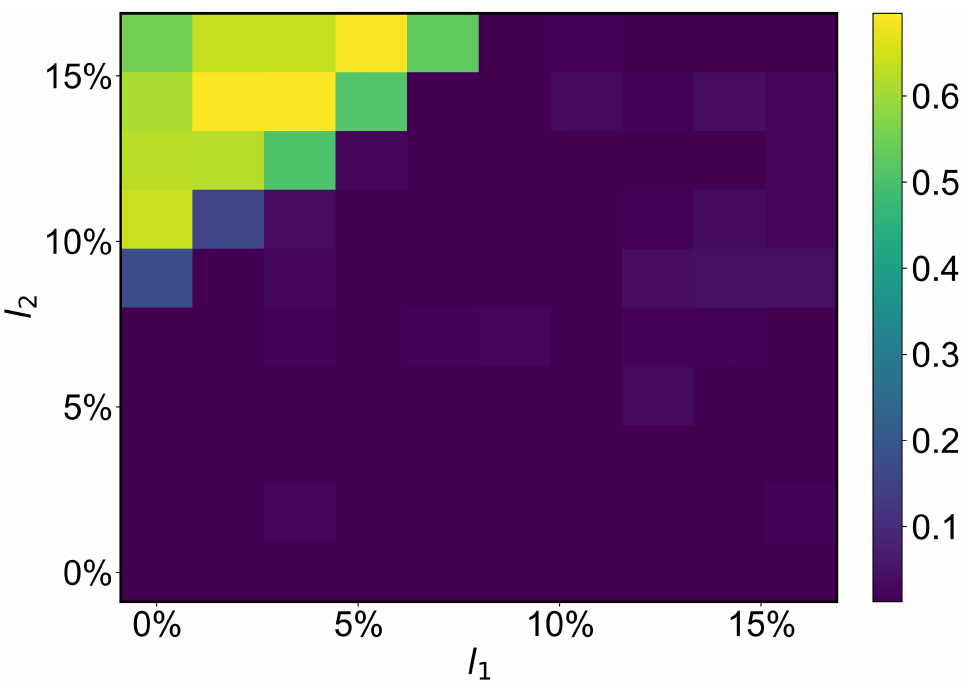} 
\caption{Robustness against parameter uncertainties in the process model. 
Shown is a color map of the testing RMSE in the parameter plane of the lengths 
$(l_1,l_2)$ of the two arms, where the chaotic Lorenz trajectory in 
Fig.~\ref{fig:4_traj}(b) is used as the reference. The ideal model parameters
used in the training are $l_1=0.5$ and $l_2=0.5$. Each RMSE value is the 
result of averaging over 50 independent realizations. The RMSE values are 
small for most of the parameter region, and the performance of the reservoir 
controller is especially robust against the uncertainty in the length of the
primary arm.}
\label{fig:uncertainty}
\end{figure}

The reservoir controller is trained for ideal parameters of the dynamical
process model. However, in applications, the parameters may differ from their 
ideal values. For example, the lengths of the two robot arms may deviate from 
what the algorithm has been trained for. More specifically, we narrow our 
attention to the uncertainty associated with the arm lengths, as variations in 
the mass parameters do not noticeably impact the control performance.
Figure~\ref{fig:uncertainty} shows the results from
the uncertainty test in tracking a chaotic Lorenz reference trajectory. It
can be seen that changes in the length $l_1$ of the primary arm have little 
effect on the performance. Only when the length $l_2$ of the secondary arm
becomes much larger than $l_1$ will the performance begin to deteriorate. 
The results suggest that our control framework is able to maintain good 
performance if the process model parameters are within reasonable limits.
In fact, when the lengths of the two robot arms are not equal, 
there are reference trajectories that the end-effector cannot physically 
track. For example, consider a circular trajectory of radius $l_1+l_2$. For 
$l_2<l_1$, it is not possible for the end effector to reach the points in the 
circle of radius $l_1-l_2$. More results from the parameter-uncertainty test
can be found in Supplementary Note 4. The issues of safe region of initial 
conditions for control success, tracking speed tolerance, and robustness
against variations in training parameters are addressed in Supplementary 
Note 5.

\section*{Discussion} \label{sec:discussion}

The two main issues in control are: (1) regularization, which involves 
designing a controller so that the corresponding closed-loop system converges 
to a steady state, and (2) tracking - to make the output of the closed-loop 
system track a given reference trajectory continuously. In both cases, the 
goal is to achieve optimal performance despite disturbances and initial 
states~\cite{TSH:book}.
The conventional method for control systems design is  
linear quadratic tracker (LQT), whose objective is, e.g., to design an optimal
tracking controller by minimizing a predefined performance index. Solutions to 
LQT in general consist of two components: a feedback term obtained by solving 
an algebraic Riccati equation and a feed-forward term which is obtained by 
solving a non-causal difference equation. These solutions require complete 
knowledge of the system dynamics and cannot be obtained in real 
time~\cite{lewis2012optimal}. Another disadvantage of LQT 
is that it can be used only for the class of reference trajectories generated 
by an asymptotically stable command generator that requires the trajectory to 
approach zero asymptotically. Furthermore, the LQT solutions are typically 
non-causal due to the necessity of backward recursion, and the infinite horizon
LQT problem is challenging in control theory~\cite{kiumarsi2014reinforcement}. 
The rapidly growing field of robotics requires the development of 
real-time, non-LQT solutions for tracking control.

We have developed a real-time nonlinear tracking control method based on 
machine learning and partial state measurements. The benchmark system employed 
to illustrate the methodology is a two-arm robotic manipulator. The goal is to 
apply appropriate control signals to make the end effector of the manipulator 
to track any complex trajectory in a 2D plane. We have exploited reservoir 
computing as the machine-learning controller. With proper training, the 
reservoir controller acquires inherent knowledge about the dynamical system 
generating the reference trajectory. 
Our inverse controller design method requires the observed state vector and
its immediate future as input to the neural network in the training phase.
The testing or deployment phase requires a combination of the current and
desired output measurements: no future measurements are needed. More
specifically, in the training phase, the input to the reservoir neural
network consists of two vectors of equal dimension: (a) the observed vector
from the robotic manipulator and (b) its immediate future version. This
design enables the controller to naturally associate the second vector with
the immediate future state of the first vector in the testing phase and to
generate control signals based on this association. After training, the 
parameters of the machine-learning controller are fixed for testing, which
distinguishes our control scheme from online learning. The controller in the 
testing phase is deployed to track a desired reference trajectory since
the immediate future vectors $\mathbf{y}(t+dt)$ are replaced by the states
generated from the desired reference trajectory, which are recognized by the 
machine as the desired immediate future states of the robotic manipulator to 
be controlled. The control signal generated in this manner compels the 
manipulator to imitate the dynamical system that generates the reference 
trajectory, resulting in precise tracking. We also take advantage of 
stochastic control signals for training the neural network to enable it to 
gain as much dynamical complexity as possible.

We have tested this reservoir computing based tracking control using a variety
of periodic and chaotic reference trajectories. In all the cases, accurate 
tracking for an arbitrarily long period of time can be achieved. We have also 
demonstrated the robustness of our control framework against input disturbance, 
measurement noise, process parameter uncertainties, and variations in 
the machine-learning parameters. A finding is that selecting the starting 
end-effector position ``wisely'' can improve the tracking success rate. In 
particular, we have introduced the concept of ``safe region'' from which the 
initial position of the end effector should be chosen (Supplementary Note 5). 
In addition, the effects of the amplitude of the stochastic control signal 
used in training and of the ``speed limit'' of the reference trajectory on the 
tracking success rate have been investigated (Supplementary Note 5). We have 
also demonstrated that feed-forward neural networks can be used to replace 
reservoir computing (Supplementary Note 6). The results suggest the practical 
utilities of our machine-learning based tracking controller: it is anticipated 
to be deployable in real-world applications such as unmanned aerial vehicle, 
soft robotics, laser cutting, soft robotics, and real-time tracking of 
high-speed air launched effects.

Finally, we remark that there are traditional methods for tracking control,
such as PID, MPC (model predictive control), and $H\infty$ trackers 
(see~\cite{xu2019generalizable,rajalakshmi2022machine}, references therein).
In terms of computational complexity, these classical
controllers are extremely efficient, while the training of our machine-learning
controller with stochastic signals can be quite demanding. However, there
is a fundamental limitation with the classic controllers: such a controller
can be effective only when its parameters were meticulously tuned for a
specific reference trajectory. For a different trajectory, a completely
different set of parameters is needed. That is, when the parameters of a
classic controller are set, in general it cannot be used to track any
alternative trajectory. In contrast, our machine-learning controller overcomes
this limitation: it possesses the remarkable capability and flexibility to
track any given trajectory after a single training session! This distinctive
attribute sets our approach apart from conventional methods, so a direct
comparison with these methods may not be meaningful. 

\section*{Methods}

\subsection*{Dynamics of joint robot arms} \label{subsec:dyn_robot_arms}

The dynamics of the system of $n$-joint robot arms can be conveniently 
described by the standard Euler-Lagrangian method~\cite{slotine1991applied}. 
Let $T$ and $U$ be the kinetic and potential energies of the system, 
respectively. The equations of motion can be determined from the system 
Lagrangian $L = T - U$ as
\begin{align}
	\frac{d}{dt} \frac{\partial L}{\partial \dot{\mathbf{q}}} - \frac{\partial L}{\partial \mathbf{q}} = \boldsymbol{\tau}, 
\end{align}
where $\mathbf{q} = [q_1,q_2,\ldots q_n]^T$ and 
$\dot{\mathbf{q}}=[\dot{q}_1,\dot{q}_2,\ldots,\dot{q}_n]^T$ are the angular
position and angular velocity vectors of the $n$ arms [with $()^T$ denoting the 
transpose], and $\boldsymbol{\tau}=[\tau_1,\tau_2,\ldots,\tau_n]^T$ is the 
external force vector with each component applied to a distinct joint denoted 
by the subscript $n$. The nonlinear dynamical equations for the robot-arm 
system can be expressed as~\cite{tang2000decentralized,hauser2011towards}
\begin{align}\label{eq:model}
	\mathcal{M}(\mathbf{q})\ddot{\mathbf{q}}+C(\mathbf{q},\dot{\mathbf{q}})\dot{\mathbf{q}} + \mathbf{G}(\mathbf{q}) + \mathbf{F}(\dot{\mathbf{q}})=\boldsymbol{\tau},
\end{align}
where $\ddot{\mathbf{q}}=[\ddot{q}_1,\ddot{q}_2,\ldots,\ddot{q}_n]^T$ is the
acceleration vector of the $n$ joints, $M(\mathbf{q})$ denotes the inertial 
matrix, $C(\mathbf{q},\dot{\mathbf{q}})\dot{\mathbf{q}}$ represents the 
Coriolis and centrifugal force, $\mathbf{G}(\mathbf{q})$ is the gravitational 
force vector, and $\mathbf{F}(\dot{\mathbf{q}})$ is the vector of the 
frictional forces at the $n$ joints which depends on the angular velocities. 
We assume that the movements of the robot arms are confined to the horizontal 
plane so that the gravitational forces can be disregarded, and we also neglect 
the frictional forces, so Eq.~\eqref{eq:model} becomes
\begin{align} \label{eq:robot_arm}
	\mathcal{M}(\mathbf{q})\ddot{\mathbf{q}}+C(\mathbf{q},\dot{\mathbf{q}})\dot{\mathbf{q}}=\boldsymbol{\tau}.
\end{align}

\begin{figure}[ht!]
\centering
\includegraphics[width=\linewidth]{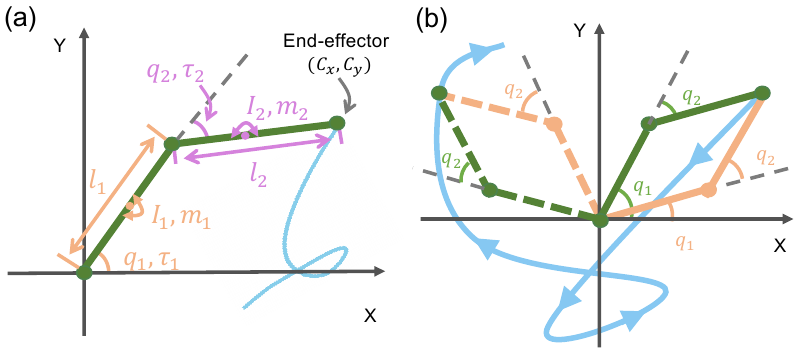}
\caption{A two-joint robot arm system and illustration of continuity of motion.
One end of the primary arm (arm 1) is fixed at the origin while the other end 
joins the secondary arm (arm 2) whose tip is the end effector.
(a) Parameter setting and dynamical variables: the moments of inertia of the 
two arms are $I_1$ and $I_2$, respectively, $m_1$ and $m_2$ are the point 
masses at the center of the two arms, the angular position of the first arm is 
$q_1$ defined with respect to the $x$ axis, and that of the second arm ($q_2$) 
is defined with respect to the direction of the first arm. The torques applied 
to the two joints are $\tau_1$ and $\tau_2$, respectively. The tip of the 
second arm is the end effector to be trained by machine learning to trace out 
any desired trajectory (the blue curve). (b) Illustration of continuity of 
motion of the two arms: two possible configurations of the arms (green and 
orange, respectively) and a trajectory (blue). For the orange configuration, 
initially the angle $q_2$ is positive because the second arm is above the line 
extending the first arm. After going through the motion as specified by the 
blue trajectory, the final angular position of the second arm is still positive.
For the green configuration, the initial angle $q_2$ is negative and it remains
to be negative after the motion. It is necessary to calculate the angles from 
Eqs.~\eqref{eq:inverse_kinematics_1} and \eqref{eq:inverse_kinematics_2} to
satisfy the continuity condition.}
\label{fig:system}
\end{figure}

We focus on the system of two joint robot arms ($n=2$), as shown in 
Fig.~\ref{fig:system}, where $m_1$ and $m_2$ are the centers of the mass of the two arms, $l_1$ and $l_2$ are their lengths, respectively. The tip of the 
second arm is the end effector to trace out a desired trajectory in the plane. 
The two matrices in Eq.~\eqref{eq:robot_arm} are 
\begin{align}
	\mathcal{M}(\mathbf{q}) &= \begin{bmatrix}
       M_{11} & M_{12} \\
       M_{21} & M_{22} \\
\end{bmatrix} \\
	\mathcal{C}(\mathbf{q},\dot{\mathbf{q}}) &= \begin{bmatrix}
		-h(\mathbf{q})\dot{q}_2 & -h(\mathbf{q})(\dot{q}_1+\dot{q}_2) \\
		h(\mathbf{q})\dot{q}_1 & 0 \\
\end{bmatrix},
\end{align}
where the matrix elements are given by 
\begin{align*}
M_{11} &=m_1 l_{\rm c_1}^2+I_1+m_2(l_1^2+l_{\rm c_2}^2+2l_1l_{\rm c_2} \cos{q_2})+I_2,\\
M_{12} &=M_{21}=m_2l_1l_{\rm c_2}\cos{q_2}+m_2l_{\rm c_2}^2+I_2,\\
M_{22} &=m_2l_{\rm c_2}^2+I_2,
\end{align*}
the function $h(\mathbf{q})$ is 
\begin{align} \nonumber
h(\mathbf{q}) =m_2l_1l_{\rm c_2}\sin{q_2},
\end{align}
$l_{\rm c_1}=l_1/2$, $l_{\rm c_2}=l_2/2$, $I_1$ and $I_2$ are the moments of inertia 
of the two arms, respectively. Typical parameter values are $m_1=m_2=1$, 
$l_1=l_2=0.5$, $l_{\rm c_1}=l_{\rm c_2}=0.25$, and $I_1=I_2=0.03$. 

The Cartesian coordinates of the end effector are 
\begin{align}
\label{eq:forward}
C_x &= l_1 \cos{q_1} + l_2 \cos(q_1+q_2),\\ \nonumber
C_y &= l_1 \sin{q_1} + l_2 \sin(q_1+q_2),
\end{align}
which give the angular positions of the two arms as
\begin{align} \label{eq:inverse_kinematics_1}
	q_2 &= \pm \arccos{\frac{C_x^2+C_y^2-l_1^2-l_2^2}{2l_1l_2}},\\ \label{eq:inverse_kinematics_2}
q_1 &= \arctan{\frac{C_y}{C_x}} \pm \arctan{\frac{l_2\sin{q_2}}{l_1+l_2\cos{q_2}}}.
\end{align}
For any end-effector position, there are two admissible solutions for the 
angular variables. We select the pair of angles that result in a continuous 
trajectory. In addition, the end effector may end up in any of the four 
quadrants, so the range of $q_1$ is $[0,2\pi]$. The range of $q_2$ is 
$[-\pi,\pi]$, since the second joint can be above or below the first joint.
In our simulations, we ensure that the solutions are continuous and thus are
physically meaningful, as demonstrated in Fig.~\ref{fig:system}(b).

Noises and unpredictable disturbances are constantly present in real-world 
applications, making it crucial to ensure that the control strategy is robust 
and operational in their presence~\cite{richard2008modern}. In fact, a model 
is always inaccurate compared with the actual physical system because of 
factors such as change of parameters, unknown time delays, measurement noise, 
and input disturbances. The goal of the robustness test is to maintain an 
acceptable level of performance under these circumstances. In our study, we 
treat disturbances and measurement noise as external inputs, where the former 
are added to the control signal and the latter is present in the sensor 
measurements. In particular, the disturbances are modeled as an additive 
stochastic process $\xi$ to the data:
\begin{align}
    \Tilde{x}_{\rm n} = x_{\rm n} + \xi_{\rm d}.
\end{align}
For measurement noise, we use multiplicative noise $\xi$ in the form 
\begin{align}
    \Tilde{x}_{\rm n} = x_{\rm n} + x_{\rm n} \cdot \xi_{\rm m}.
\end{align}
Both stochastic processes $\xi_{\rm d}$ and $\xi_{\rm m}$ follow a normal distribution of
zero mean and with standard deviation $\sigma_{\rm d}$ and $\sigma_{\rm m}$, respectively.

\subsection*{Inverse design based controller formulation} \label{subsec:control_formulation}

To develop a machine-learning based control method, it is necessary to obtain 
the control signal through observable states. The state of the two-arm system, i.e., 
the dynamical process to be controlled, is eight-dimensional, which consists of 
the Cartesian coordinates of the end-effector, the angular positions, angular 
velocities and angular accelerations of the two manipulators: 
\begin{align} \label{eq:full_state} 
\mathbf{x}\equiv [C_x,C_y,q_1,q_2,\dot{q}_1,\dot{q}_2,\ddot{q}_1,\ddot{q}_2]^T.
\end{align}
A general nonlinear control problem can be formulated as~\cite{canaday2021model}
\begin{align} \label{eq:plant}
\mathbf{x}(t+dt)&=\mathbf{f}[\mathbf{x}(t),\mathbf{u}+\mathbf{u}\cdot\xi_{\rm d}], \\
\label{eq:contorller}
\mathbf{y}(t)&=\mathbf{g}[\mathbf{x}(t)]+\mathbf{g}[\mathbf{x}(t)]\cdot\xi_{\rm m},
\end{align}
where $\mathbf{x}\in \mathbb{R}^n$ ($n=8$), $\mathbf{u}\in \mathbb{R}^m$ 
($m<n$) is the control signal, $\mathbf{y}\in \mathbb{R}^k$ ($k\leq n$) 
represents the sensor measurement. The function $\mathbf{f}: 
\mathbb{R}^n \times \mathbb{R}^m \rightarrow \mathbb{R}^n$ is unknown for 
the controller. In our analysis, we assume that $\mathbf{f}$ is Lipschitz 
continuous~\cite{o2006metric} with respect to $\mathbf{x}$. The measurement
function $\mathbf{g}: \mathbb{R}^n \rightarrow \mathbb{R}^k$ fully or partially
measures the states $\mathbf{x}$. For the two-arm system, the measurement 
vector is chosen to be four-dimensional: 
$\mathbf{y} \equiv [C_x, C_y, \dot{q_1}, \dot{q_2}]^T$. The corresponding 
vector from the desired, reference trajectory is denoted as 
$\mathbf{y}_{\rm d}(t)$. For our tracking control problem, the aim is to design
a two-degree-of-freedom controller that receives the signals $\mathbf{y}(t)$ 
and $\mathbf{y}_{\rm d}(t)$ as the input and generates an appropriate control 
signal $\mathbf{u}(t)$ in order for $\mathbf{y}(t)$ to track the trajectory
generating the observation $\mathbf{y}_{\rm d}(t)$. For convenience, we use 
the notation $\mathbf{f}_{\rm u}(\cdot) \equiv \mathbf{f}(\cdot,\mathbf{u})$.
For a small time step $dt$, Eq.~\eqref{eq:plant} becomes
\begin{align} \label{eq:inversC_x}
\mathbf{x}(t+dt) \approx \mathbf{F}_{\rm u}[\mathbf{x}(t)],
\end{align}
where $\mathbf{F}_{\rm u}$ is a nonlinear function mapping $\mathbf{x}(t)$ 
to $\mathbf{x}(t+dt)$ under the control signal $\mathbf{u}(t)$. For reachable
desired state, $\mathbf{F}_{\rm u}$ is invertible. We get
\begin{align}
\label{eq:inverse_f}
\mathbf{u}(t) \approx \mathbf{F}_{\rm u}^{-1}[\mathbf{x}(t),\mathbf{x}(t+dt)],
\end{align}
Similarly, Eq.~\eqref{eq:contorller} can be approximated as 
$\mathbf{x}(t) \approx \mathbf{g}^{-1}[\mathbf{y}(t)]$, so 
Eq.~\eqref{eq:inverse_f} becomes
\begin{align} \label{eq:u}
\mathbf{u}(t) \approx \mathbf{F}^{-1}[\mathbf{g}^{-1}[\mathbf{y}(t)],\mathbf{g}^{-1}[\mathbf{y}(t+dt)]].
\end{align}
Equation~\eqref{eq:u} is referred to as the inverse model for nonlinear 
control~\cite{canaday2021model}, which will be realized in a model-free 
manner using machine learning.

\section*{Data Availability}
\vspace*{-0.1in}

The reference trajectories data generated in this study can be found in the 
repository: https://doi.org/10.5281/zenodo.8044994~\cite{zhai:data}.

\section*{Code Availability}
\vspace*{-0.1in}
The codes for generating all the results can be found on GitHub: 
https://github.com/Zheng-Meng/TrackingControl~\cite{zhai:code}.

\newpage

%\bibliographystyle{naturemag}
%\bibliography{COC}

\section*{Acknowledgments}
\vspace*{-0.1in}
This work was supported by the Army Research Office through Grant 
No.~W911NF-21-2-0055 (to Y.-C.L.) and by the Air Force Office of Scientific
Research under Grant No.~FA9550-21-1-0438 (to Y.-C.L.).

\section*{Author Contributions}
\vspace*{-0.1in}
Z.-M. Z., M. M, L.-W. K., B. G., M. H., and Y.-C. L. designed the research 
project, the models, and methods. Z.-M. Z. performed the computations. 
Z.-M. Z., M. M, L.-W. K., B. G., M. H., and Y.-C. L. analyzed the data. 
Z.-M. Z. and Y.-C. L. wrote the paper. M. H. and Y.-C. L. edited the manuscript.

\section*{Competing Interests}
\vspace*{-0.1in}
The authors declare no competing interests.

\end{document}